\pdfoutput=1

\documentclass[11pt]{article}

\usepackage[]{acl2023}

\usepackage{times}
\usepackage{latexsym}

\usepackage[T1]{fontenc}

\usepackage[utf8]{inputenc}

\usepackage{microtype}

\usepackage{inconsolata}
\usepackage{soul}
\usepackage{linguex}
\usepackage{graphicx}
\usepackage{url}
\usepackage{amssymb}
\usepackage{subcaption}
\usepackage{multirow}
\usepackage{kotex}
\usepackage{mdframed}

\title{Do language models capture implied discourse meanings? \\An investigation with exhaustivity implicatures of Korean morphology}

\author{Hagyeong Shin \\
  Department of Linguistics \\
  University of California San Diego \\
  \texttt{hashin@ucsd.edu} \\\And
  Sean Trott \\
  Department of Cognitive Science\\  University of California San Diego\\ 
  \texttt{sttrott@ucsd.edu} \\}

\begin{document}
\maketitle
\begin{abstract}

Markedness in natural language is often associated with non-literal meanings in discourse. 
Differential Object Marking (DOM) in Korean is one instance of this phenomenon, where post-positional markers are selected based on both the semantic features of the noun phrases and the discourse features that are orthogonal to the semantic features. Previous work has shown that distributional models of language recover certain semantic features of words---do these models capture implied discourse-level meanings as well?
We evaluate whether a set of large language models are capable of associating discourse meanings with different object markings in Korean. 
Results suggest that discourse meanings of a grammatical marker can  be more challenging to encode than that of a discourse marker.

\end{abstract}

\section{Introduction}
The distributional hypothesis states that the meaning of a word can be derived in part from the linguistic contexts in which it is used \cite{harris_distributional_1954}. Accordingly, language models built on distributional patterns of language use have proved useful for modeling both content words \cite{lenci_distributional_2018, boleda_distributional_2020} and function words \cite{marelli_affixation_2015}. With recent advancements, large scale neural language models have demonstrated that distributional semantic models encode substantial amount of semantic knowledge \cite{tenney_bert_2019, trott_raw-c_2021} and even contextualized meanings of words \cite{tenney_what_2019}. In the current work, we ask whether distributional semantics in a large scale can encode meanings of function words that encompasses semantic and discourse meanings.

Differential Object Marking (DOM), particularly in Korean, poses an interesting challenge to distributional semantics. DOM
is a phenomenon in which grammatical objects can be marked with different grammatical structures (e.g., case markers) \cite[see][]{wanner_differential_1991}, often explained with varying semantic features of the referent \cite[see][]{aissen_differential_2003}. In Korean, DOM is associated not only with the \textit{semantic} features of an object but also with the \textit{discourse status} of the object \cite{kwon_differential_2008, lee_iconicity_2006}.  
As an instance, three alternative markings---\textit{lul}, \textit{nun},\footnote{The \textit{lul} and \textit{nun} markers are allomorphic and are respectively realized as \textit{ul} and \textit{un} to follow a syllable that ends with a coda (consonant). For clarity, we refer to each marker as \textit{lul} and \textit{nun}.} and null-marking---can appear after an object that is picked out from a set of contextualized alternatives. While the null-marking option reflects that the object is contextualized element in the discourse context, \textit{lul} and \textit{nun} markers imply the \textit{exhaustive status} of the object, contrasted with other alternatives from the set.
Furthermore, each of the \textit{lul} and \textit{nun} markers derives the exhaustive status that posit different constraint on the upcoming discourse.

Thus, in order to account for Korean DOM, distributional patterns of object markings must be able to encode multiple meanings of the markers at once: the discourse features that are grounded with the discourse context, as well as semantic features of objects that are orthogonal to discourse context. 
To test this, we used pre-trained Large Language Models (LLMs).
LLMs have been evaluated as a subject of psycholinguistics \cite{futrell_neural_2019} for their ability to grasp non-literal meanings \cite{hu_fine-grained_2023, jeretic_are_2020} and those that are closely tied to the pragmatic context \cite{trott_large_2023}. In line with these studies, we evaluate whether LLMs exhibit the competency with  discourse meanings associated with different object markings in Korean.

\section{Patterns under evaluation}

In this study, we focus on three post-positional marking options in Korean, as illustrated in \Next.\footnote{Following abbreviations are used in the examples. $\emptyset$ = null-marking, \textsc{acc} = accusative, \textsc{add} = additive, \textsc{ct} = contrastive, \textsc{nom} = nominative, \textsc{neg} = negation, \# = infelicitous.} The first \textit{lul} marker is canonically an accusative case marker. Without any considerations on specific discourse context, the \textit{lul} marker indicates that a noun phrase is a grammatical direct object. The \textit{nun} marker, on the other hand, is mainly used as a discourse marker and replaces the \textit{lul} marker. In \Next, the \textit{nun} marker indicates that the object is being contrasted with other entities \cite{choi_optimizing_1996,kim_deriving_2018}. The last marking option in \Next, annotated with $\emptyset$, indicates that no markers follow the object. This option, what we refer to as ``null-marking'', is common in colloquial Korean \cite{choi-jonin_particles_2008, lee_case_2007}. All of the three marking options in \Next are grammatical and felicitous to appear after an object.

\exg. Mina-ka pizza-\textbf{\{lul/nun/$\emptyset$\}} kacyewasse.\\
Mina-\textsc{nom} pizza-\{\textsc{acc}/\textsc{ct}/$\emptyset$\} brought\\
`Mina brought a pizza.'

Some options listed in \Last, however, can lead to  \textit{implicatures} in a discourse context.
Implicatures refer to the meaning that is not conveyed by the truth-conditional meaning of each words, but what can be inferred from those in the context of the usage \cite{grice_studies_1989}.
As a specific type of implicature, \textit{exhaustivity implicature} arises when there is a set of alternatives in the discourse. When one element from the set is picked out as an answer to the question, the mentioned element is perceived as an exhaustive information relevant to the question \cite{buring_d-trees_2003, horn_exhaustiveness_1981, rooth_theory_1992, van_rooij_exhaustive_2004}.

The \textit{lul} and \textit{nun} marker can both evoke exhaustivity implicatures, while the null-marking does not.
To illustrate, in \Next, the given context evokes a set of alternatives \{\textit{pizza}, \textit{cake}\}. In the first response option (A), one alternative \textit{pizza} from the set is picked out but appears without any marker. In this response, the null-marking does not convey an exhaustivity status of the object, but indicates that the speaker refers to the contextualized object in the discourse context.

\ex. Context: Interlocutors know that guests were invited to bring a pizza and a cake to the party, and that Mina attended the party.
\a.[Q. ] What did Mina bring?
\bg.[A. ] Pizza-$\mathbf{\emptyset}$ kacyewasse.\\
pizza-$\emptyset$ brought \\
`(Mina) brought the pizza.' 
\bg.[B. ] Pizza-\textbf{lul} kacyewasse. Cake-to kacyewasse.\\
pizza-\textsc{acc} brought cake-\textsc{add} brought\\
`(Mina) brought the pizza.\\
($\rightarrow$ \st{Mina didn't bring anything else.}) (Mina) also brought the cake.' 
\bg.[C. ] Pizza-\textbf{nun} kacyewasse. \#Cake-to kacyewasse.\\
pizza-\textsc{ct} brought cake-\textsc{add} brought\\
`(Mina) brought the pizza.\\
($\rightarrow$ Mina didn't bring anything else.) (Mina) also brought the cake.' 
\bg.[C$'$. ] Pizza-\textbf{nun} kacyewasse. Cake-un ahn kacyewasse.\\
pizza-\textsc{ct} brought cake-\textsc{ct} \textsc{neg} brought\\
`(Mina) brought the pizza.\\
($\rightarrow$ Mina didn't bring anything else.) (Mina) didn't bring the cake.'

Both of the \textit{lul} and \textit{nun} markers can evoke exhaustivity implicatures, but each marker's implicatures differ in terms of whether the information conveyed by the implicature can be corrected in the upcoming discourse. In other words, the \textit{lul} and \textit{nun} marker's exhaustivity implicatures exhibit different \textit{cancelability} \cite{lee_contrastive_2003,lee_contrastiveness_2017}.
In \Last B, \textit{pizza} is picked out from the set of alternatives and marked with the \textit{lul}. The \textit{lul} marker conveys an exhaustivity implicature (given in the parentheses), which is cancelable in the upcoming discourse \cite{lee_contrastive_2003}. Due to its cancelability, the second utterance `(Mina) also brought a cake.' forms a felicitous continuation of the response.

C and C$'$ in \Last illustrates the exhaustivity implicature evoked by the \textit{nun} marker. However, unlike the \textit{lul} marker, the \textit{nun} marker evokes an exhaustivity implicature that is not cancelable \cite{kim_deriving_2018, lee_contrastive_2003}. In C, the second sentence cannot be the felicitous discourse continuation, because it contradicts the uncancelable exhaustivity implicature derived by the \textit{nun} marker. C$'$ presents the felicitous discourse continuation, where the exhaustivity status of the object indicated by the \textit{nun} is not contradicted.\footnote{To be more precise, exhaustive interpretations of the \textit{lul} and \textit{nun} markers have been discussed in association with different types of Information Structure \cite{lambrecht_information_1994}. The \textit{lul} marker is associated with the Contrastive Focus (CF) status and derives the exhaustivity of the CF element(s) \cite{lee_contrastive_2003, lee_contrastiveness_2017}. The \textit{nun} marker is associated with the Contrastive Topic (CT) status, and implies that there is unanswered portion of a Question Under Discussion (QUD) \cite{buring_d-trees_2003}. To identify discourse patterns that language models can be evaluated for, we defer discussions on theoretical notions.}

Thus, paradigmatic contrasts between \textit{lul}, \textit{nun}, and null-marking derive non-literal meanings that are captured in the discourse domain: exhaustivity implicatures and their cancelability in association with each marking options.
In the following sections, we examine whether LLMs' semantic representations of each marking options reflect these discourse meanings.

\section{Models under evaluation}

To assess distributional semantics with discourse meanings of Korean DOM, we examine a set of pre-trained LLMs. Pre-trained LLMs are trained to perform a token prediction task on large volumes of corpora (e.g., sometimes billions or trillions of words) using many parameters. LLMs learn to predict words in context by observing statistical patterns in which words and word sequences are most likely to co-occur. This makes them well-suited as operationalizations of the distributional hypothesis. 

Previous research suggests that distributional semantics may be able to effectively encode discourse meanings, particularly when the training data and the model are both sufficiently large.
\citet{tenney_what_2019} discovered that more contextualized knowledge may emerge in deeper layers of a model's architecture.
\citet{jeretic_are_2020} found that models are adept at learning non-literal meanings, even though off-the-shelf models may lack pragmatic competency. 
Additionally, \citet{hu_fine-grained_2023} observed that larger models achieve high accuracy and align with human error patterns in various pragmatic phenomena. 
Considering these results, we evaluate models with different numbers of parameters to determine whether distributional semantics can effectively encode discourse meanings associated with different object markings as the models scale up.

We test a series of generative pre-trained transformer models that are developed for the Korean language and as multi-lingual models. Starting from models with smaller parameters to larger ones, we include KoGPT-2 (125M)\footnote{\url{https://huggingface.co/skt/kogpt2-base-v2}} and KoGPT-Trinity (1.2B),\footnote{\url{https://huggingface.co/skt/ko-gpt-trinity-1.2B-v0.5}} both developed and trained specifically for Korean.
We also test Polyglot-Ko models with 3.8B,\footnote{\url{https://huggingface.co/EleutherAI/polyglot-ko-3.8b}} 5.8B,\footnote{\url{https://huggingface.co/EleutherAI/polyglot-ko-5.8b}} and 12.8B\footnote{\url{https://huggingface.co/EleutherAI/polyglot-ko-12.8b}} parameters, developed under a project for multilingual LLMs and primarily trained with Korean data.
Lastly, we test text-davinci-003/GPT-3 (175B)\footnote{Accessed before its deprecation on January 4th, 2024.} and gpt4-1106-preview/ChatGPT, both accessed with OpenAI API.\footnote{\url{https://platform.openai.com/docs/api-reference}} These two models stand out as they are trained with reinforcement learning from human feedback (RLHF); this makes them less suitable to a direct test of the ``pure'' distributional hypothesis (given that they receive explicit human feedback), but they remain useful operationalizations of what can be learned from a linguistic training signal. Despite not being developed specifically for Korean, their massive size and diverse training data enable them to demonstrate competency in using the Korean language.

All models, with the exception of ChatGPT, grant access to their internal semantic representations through logits/log probabilities assigned to input and output tokens. ChatGPT, on the other hand, is optimized for `prompting,' limiting the assessment of its semantic representation via the model's meta-judgments on inputs. We employ distinct approaches to evaluate these two model types.

\section{Experiment 1: Discourse meanings in processing}

We first investigate whether LLMs' semantic representation of words can reflect discourse meanings of the \textit{lul} and \textit{nun}.  We compare LLMs' and humans' sensitivity towards exhaustivity implicatures indicated with the \textit{lul} and \textit{nun} markers and the different cancelability of each marker's implicatures.

\begin{table}[ht]
    \centering
    \begin{tabular}{|p{1.8cm}p{5cm}|}
    \hline
        \textit{Set of alternatives}  & Sohee knows that Mina and Yuna could receive medals and trophies.\\  
        \textit{Question} & Sohee: What did Yuna receive? \\ 
        \textit{Object marking} & Mina: Received a medal-\textbf{\{lul/ nun\}}. \\ 
        \textit{Continuation} & \textbf{\{Also received/Didn't receive/Only received\}} a trophy.\\
        \hline      
    \end{tabular}
    \captionof{table}{An illustration of items used in the Experiment 1. Non-italic components in the right column are presented in Korean, and only one option in curly brackets is presented in a single item. See the Appendix \ref{sec:appendixA} for this example written in Korean.
    }
    \label{tab:exp1items}
  \end{table}

\begin{figure*}[ht!]
    \centering
    \begin{subfigure}[t]{0.65\textwidth}
        \centering
    \includegraphics[scale=0.55]{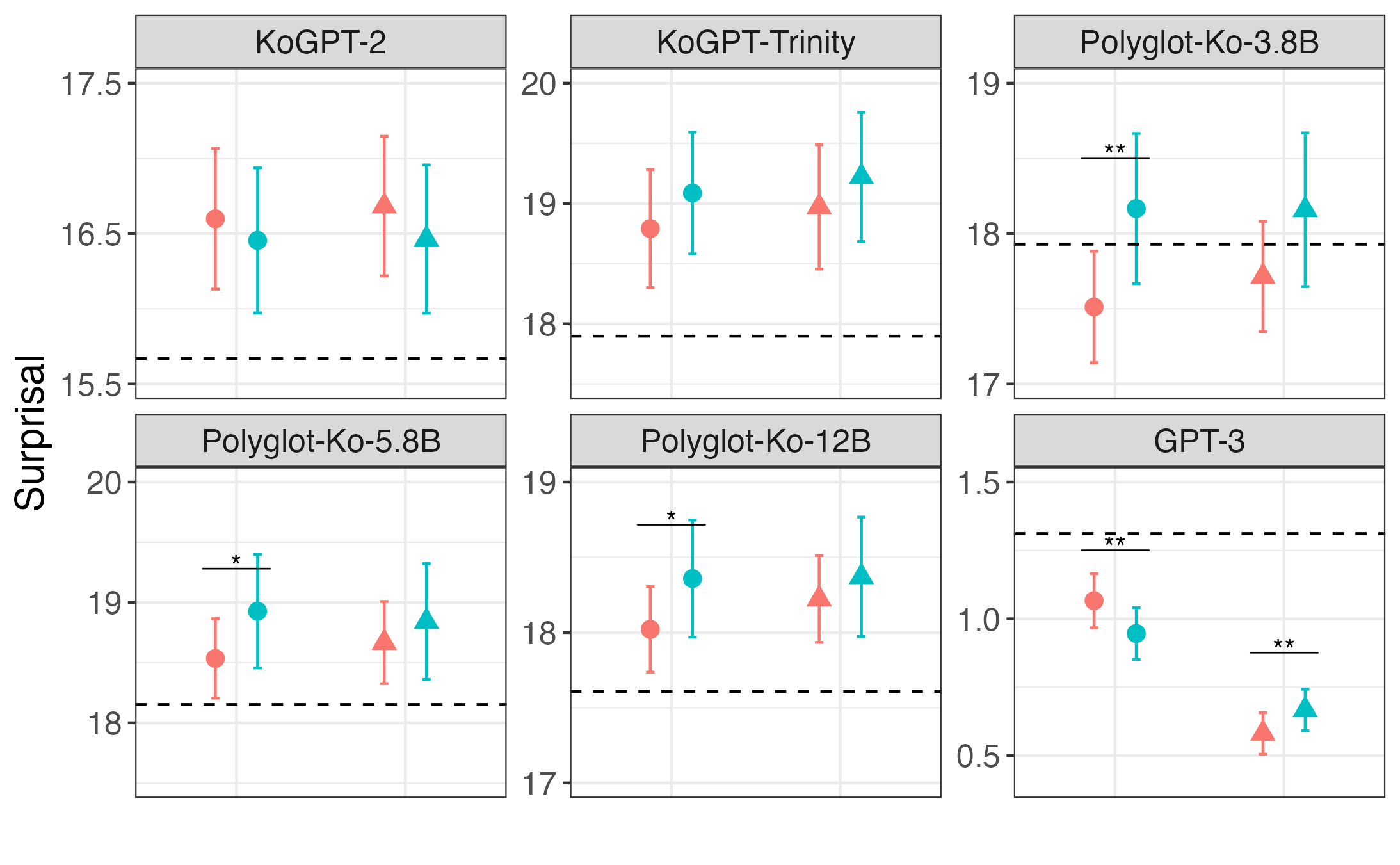}
    \end{subfigure}~ 
    \begin{subfigure}[t]{0.35\textwidth}
        \centering        \includegraphics[scale=0.55]{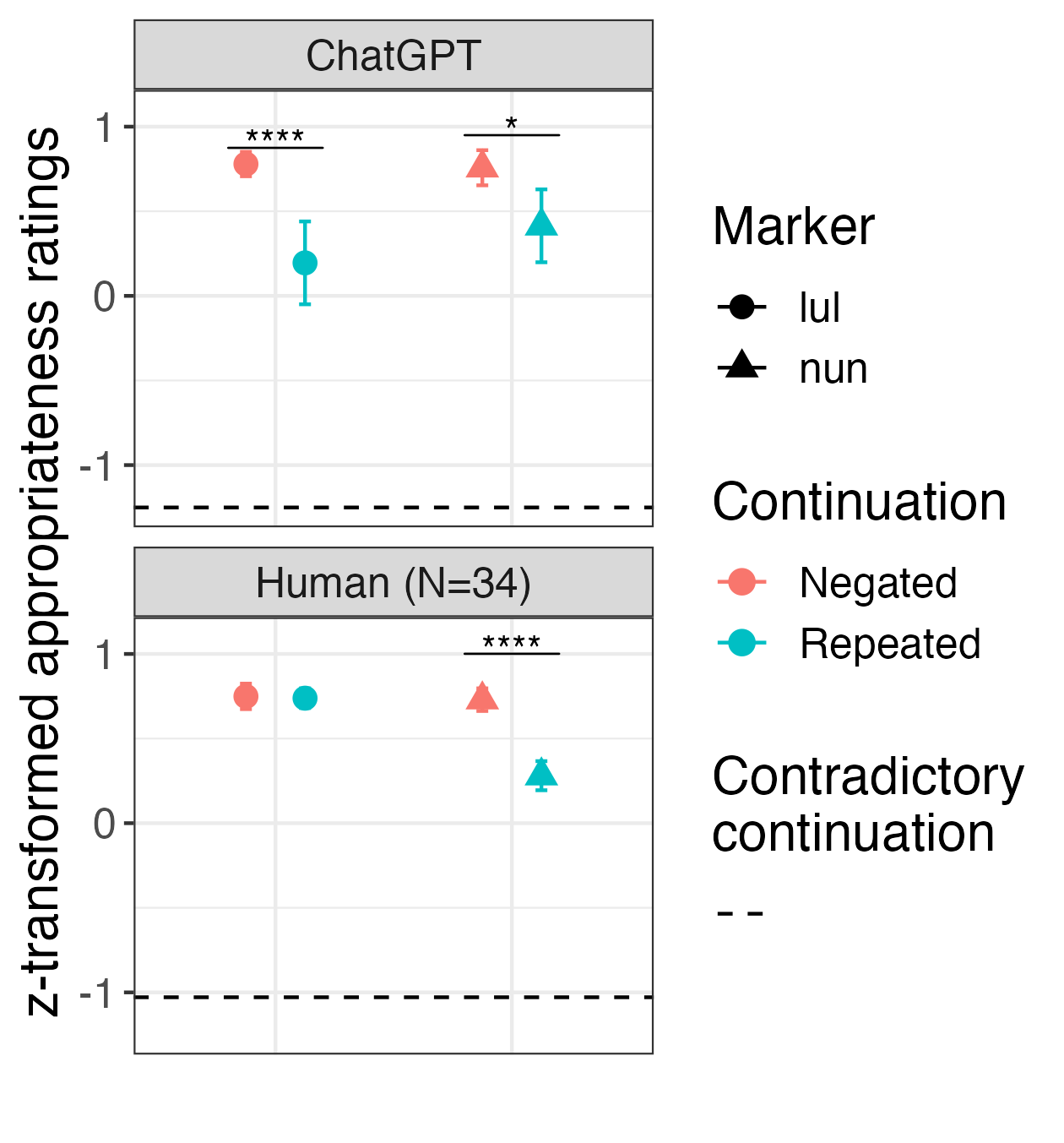}
\end{subfigure}
    \caption{Evaluations of discourse continuations where the exhaustivity status implied by \textit{lul} and \textit{nun} is canceled (with repeated continuations) or not canceled (with negated continuations), and where the previous sentence is logically contradicted (dashed line). Mean surprisals and 95\% CIs gathered from the 6 models are presented on the left. Higher surprisals indicate that the model had lower expectations for encountering the continuation. On the right side, mean of z-transformed appropriateness ratings and 95\% CIs from ChatGPT and 34 native Korean speakers are presented. Higher ratings indicate that the discourse continuation was evaluated as more felicitous. Stars indicate adjusted significance levels obtained from paired $t$-tests with Bonferroni corrections (****: $p < 0.001$, **: $p < 0.005$, *: $p < 0.05$).}
    \label{fig:processing}
\end{figure*}

We created 288 stimuli (48 items to appear in 6 conditions), as shown in Table \ref{tab:exp1items}. Each item begins with a sentence contextualizing a set of alternatives, followed by a question about one alternative from the set. The response portion of the conversation always consists of two sentences. The first sentence was manipulated to have different object markings between \textit{lul} and \textit{nun}. The second sentence states how the other alternative from the set forms the relation with the elided subject. The previous verb from the question (and the first response sentence) is manipulated to be either repeated or negated. When the previous verb is repeated, the exhaustivity status established in the previous sentence is canceled; when it is negated, the exhaustivity status is maintained. As a baseline, we also included contradictory continuations where the alternative object (e.g., \textit{trophy} in Table \ref{tab:exp1items}) is marked with the `only' (-\textit{man}) marker and evokes a logical contradiction with the first sentence.

\subsection{Surprisal measurements}

As an assessment of semantic representations of different marking options, we obtained surprisal of each sentence following different markings (e.g., \textit{Continuation} sentence in Table \ref{tab:exp1items}) from all models except ChatGPT. Logits assigned to each token in the sentence are first converted into log probabilities. We then summed log probabilities of all tokens in a sentence and normalized for the number of tokens in a sentence. These are then converted to surprisal. For GPT-3, we skipped the step of converting logits into log probabilities, as the output already provides log probabilities.
If models perceive the exhaustivity implicature and its cancelability, they should exhibit notably higher surprisal only when the \textit{nun} marker is followed by a repeated verb continuation (e.g., ``also received'' in Table \ref{tab:exp1items}).

Surprisal scores are summarized in the left subfigure of Figure \ref{fig:processing}. 
Among the six models, only the GPT-3 could distinguish logical contradiction (dashed line in Figure \ref{fig:processing}) from pragmatically infelicitous discourse. Additionally, GPT-3 was the only model sensitive to the non-cancelable exhaustivity implicature of the \textit{nun} marker.
However, the model exhibited higher surprisal when encountering an uncanceled implicature following the \textit{lul} marker, deviating from the observed pattern in \Last B and from human speakers. All Polyglot-Ko models showed sensitivity towards the \textit{lul} marker followed by non-canceled exhaustivity, which is also incoherent with the cancelability observed in \Last B and from human speakers.

\begin{table*}[ht]
    \centering
    \begin{small}
    \begin{tabular}{l|cccccc|cc} \hline
        \multirow{3}{*}{Coefficients}& \multicolumn{6}{c|}{$\beta$ of surprisal } & \multicolumn{2}{c}{$\beta$ of z-transformed}\\
        & \multicolumn{6}{c|}{($p$-value) } & \multicolumn{2}{c}{ratings ($p$-value)} \\
        \cline{2-9}
        & KoGPT-2 & Ko-Trinity & Poly-3B & Poly-5B & Poly-12B & GPT-3 & ChatGPT & Human \\ \hline 
    \multirow{2}{*}{Intercept} & 16.60  & 18.79  & 17.51& 18.53 & 18.02   & 1.07  & 0.78 &  0.75\\
    & (***) & (***) & (***) & (***) & (***)  &(***) &(***) &(***) \\ 
    \multirow{2}{*}{Marker:Nun} & 0.08  & 0.18 & 0.20 & 0.13 & 0.20   & -0.49  &-0.02 &-0.02 \\
    & (0.46) & (0.30) & (0.15)& (0.24) & (*)&  (***) & (0.85)& (0.65) \\
    \multirow{2}{*}{Continuation:Repeated} & -1.14  & 0.30  & 0.65 & 0.39 & 0.34 & -0.12  & -0.58 &  
    -0.01 \\
    & (0.21) & (0.09) & (***)& (***) &(***)  &(***)& (***) &(0.88) \\
    Marker:Nun$\times$
    & -0.08 & -0.05 & -0.21 & -0.22 &-0.19  & \textbf{0.21}  &0.24 &\textbf{-0.44} \\ 
    Continuation:Repeated & (0.64) &  (0.85)& (0.29)& (0.17) &(0.14)  &\textbf{(***) }& (0.15)& \textbf{(***)}\\ \hline
    \end{tabular} \end{small}
    \caption{Surprisal scores obtained with each of the first six models are fitted with the mixed effects model: \texttt{surprisal $\sim$ marker * continuation + (1|item)}. Positive coefficients indicate increase in the surprisal, thus decrease in the model's expectedness. Ratings obtained with ChatGPT and human are z-transformed and fitted with the mixed effects model. For ChatGPT: \texttt{z-rating $\sim$ marker * continuation + (1|item)}. For human: \texttt{z-rating $\sim$ marker * continuation + (1+marker*continuation|participant) + (1+marker*continuation|item)}. In these models, positive coefficients indicate increase in the ratings of appropriateness. ***: $p < 0.001$, *: $p<0.05$.}
    \label{tab:processing-regression}
\end{table*}

\subsection{Elicited ratings}

ChatGPT's and humans' processing patterns were assessed by asking them to rate how much each type of continuations (c.f. Table \ref{tab:exp1items}) is appropriate to follow the previous sentence. A 7-point likert scale was provided, with 1 representing the second sentence being `inappropriate' and 7 representing it being `appropriate'. For ChatGPT, we set a system message---\textit{Make sure to respond only with a number between 1 and 7.}---via OpenAI API. This can be considered as a meta-instruction guiding how the model should respond to each prompt. We also set the temperature to 0, which guides the model to generate responses deterministically based on the probability assignments. For human participants, the context preamble was written more specifically to describe the shared knowledge of speakers and listeners regarding the set of alternatives (see the Appendix \ref{sec:appendixA} for the full item). In the human experiment, contradictory continuations were presented as a part of fillers. Each participant saw 24 critical items appearing in one out of 4 conditions, 24 fillers, and 4 attention checks. Native speakers of Korean were recruited and compensated via online crowdsourcing platform based in South Korea.\footnote{\url{https://gosurveasy.com/}} After the data elimination process, responses from 34 participants are retained and reported.

We transformed ChatGPT's and each of the 34 participant's appropriateness ratings, including ratings on contradictory continuations, into z-scores, using the mean and standard deviation obtained within each participant/model.
The z-transformed ratings are presented in the right subfigure in Figure \ref{fig:processing}. Raw ratings without z-transformation are presented in the Appendix \ref{sec:appendixB}.
When interpreting ChatGPT's results, we adopt a cautious and non-conclusive approach, guided by the findings of \citet{hu_prompt-based_2023}. Instead of viewing the chat responses as a direct reflection of the model's semantic representation, we consider them as indicative of the model's proficiency in making evaluations about the input.

In general, ChatGPT and humans showed positive ratings for critical items targeting appropriateness ratings, while giving negative ratings for contradictory continuations targeting truth-conditional judgements (dashed line in Figure \ref{fig:processing}). This confirms that their ratings were based on discourse (in)felicity, not on truth-conditional meanings.
ChatGPT showed similar sensitivity observed with GPT-3. The model was somewhat sensitive towards the \textit{nun} marker's non-cancelable exhaustivity implicature, which was coherent with human speakers' patterns. However, the model also produced more negative ratings for the \textit{lul} marker's cancelable implicature, which differed from the pattern observed in human speakers.

\subsection{Mixed-effects models}

We fitted a mixed effects regression model to further investigate each model's sensitivity towards the cancelability of the exhaustivity implicature. 
For models assessed with surprisal measurements, we fitted a model predicting surprisal with marker, continuation, and the interaction of the two, while controlling for random effects of each lexicalized item. For ChatGPT and humans assessed with ratings, we fitted a model predicting z-transformed ratings. For predicting humans' ratings, random effects of each participant were additionally controlled.  Results are summarized in Table \ref{tab:processing-regression}, with models listed from smaller to larger ones.

Comparing the first two KoGPT series model with Polyglot-Ko models, we observe that the Polyglot-Ko models have gained sensitivity towards the repeated continuation. As the Polyglot-Ko models reaches 12B parameters, they also gain sensitivity towards the markedness of an object. 
GPT-3 exhibited sensitivity to the infelicity arising from the \textit{nun} marker followed by canceled exhaustivity (Marker:Nun$\times$Continuation:Repeated). Additionally, it displayed sensitivity to other factors that human speakers did not perceive as influencing discourse felicity.
ChatGPT demonstrated sensitivity only to the continuation factor that human speakers were insensitive to.

\section{Experiment 2: Discourse meanings in production}

From the previous experiment, we observed that Polyglot-Ko models and GPT-3 were surprised to see the \textit{lul} marker followed by the canceled exhaustivity, while human speakers accepted the \textit{lul} marker followed by either canceled or non-canceled exhaustivity. As noted with the example \LLast in Section 2, the \textit{lul} marker is canonically a grammatical case marker, and its grammatical function persists without the considerations on discourse context. Although the experimental items were designed to evoke the \textit{lul} marker's discourse function, it needs to be confirmed that the results from Experiment 1 stem from models and humans associating the \textit{lul} marker with its additional discourse meaning (exhaustivity implicature), not solely from its grammatical function (marking the grammatical objects). To address this, we conducted the second experiment.

\begin{table}[b]
    \centering
    \begin{tabular}{|p{1.5cm}p{5.3cm}|} \hline
        \textit{Intended message} &  Mina intends to respond that Yuna received \textbf{\{only the medal/both the medal and the trophy\}}. \\ 
        \textit{Question} & Sohee: What did Yuna receive? \\
        \textit{Response} & Medal-\textbf{\{lul/nun/$\emptyset$\}} received.\\ \hline
    \end{tabular}
    \caption{An example of items used in the Experiment 2. Non-italic components in the right column are presented in Korean, and only one option among others in curly brackets are presented in the actual item. See the Appendix \ref{sec:appendixC} for the item written in Korean.}
    \label{tab:production_item}
\end{table}

We designed 480 stimuli (48 items to appear with 10 manipulated components) exemplified in Table \ref{tab:production_item}. Each item started with the \textit{intended message}, wherein the exhaustivity status of an object that a speaker intends to indicate was manipulated. Then, a question on the object and the response with three different marking options followed. The response sentence could have \textit{lul}-, \textit{nun}-, or null-marked object. As a baseline, we included a `contradictory' response that does not match the (non-)exhaustivity status described in the intended message. Additionally, a `verbatim repeat' response was included, maintaining the exact structure of the relative clause from the intended message statement. If LLMs can indeed associate the \textit{lul} and \textit{nun} markers with the exhaustivity status of the object, they should generate \textit{lul}-marked and \textit{nun}-marked responses when exhaustivity of the object is intended, and null-marked responses when exhaustivity of the object is not intended.

\begin{figure*}[ht!]
    \centering
    \begin{subfigure}[t]{0.49\textwidth}
        \raggedleft
    \includegraphics[width=0.65\textwidth]{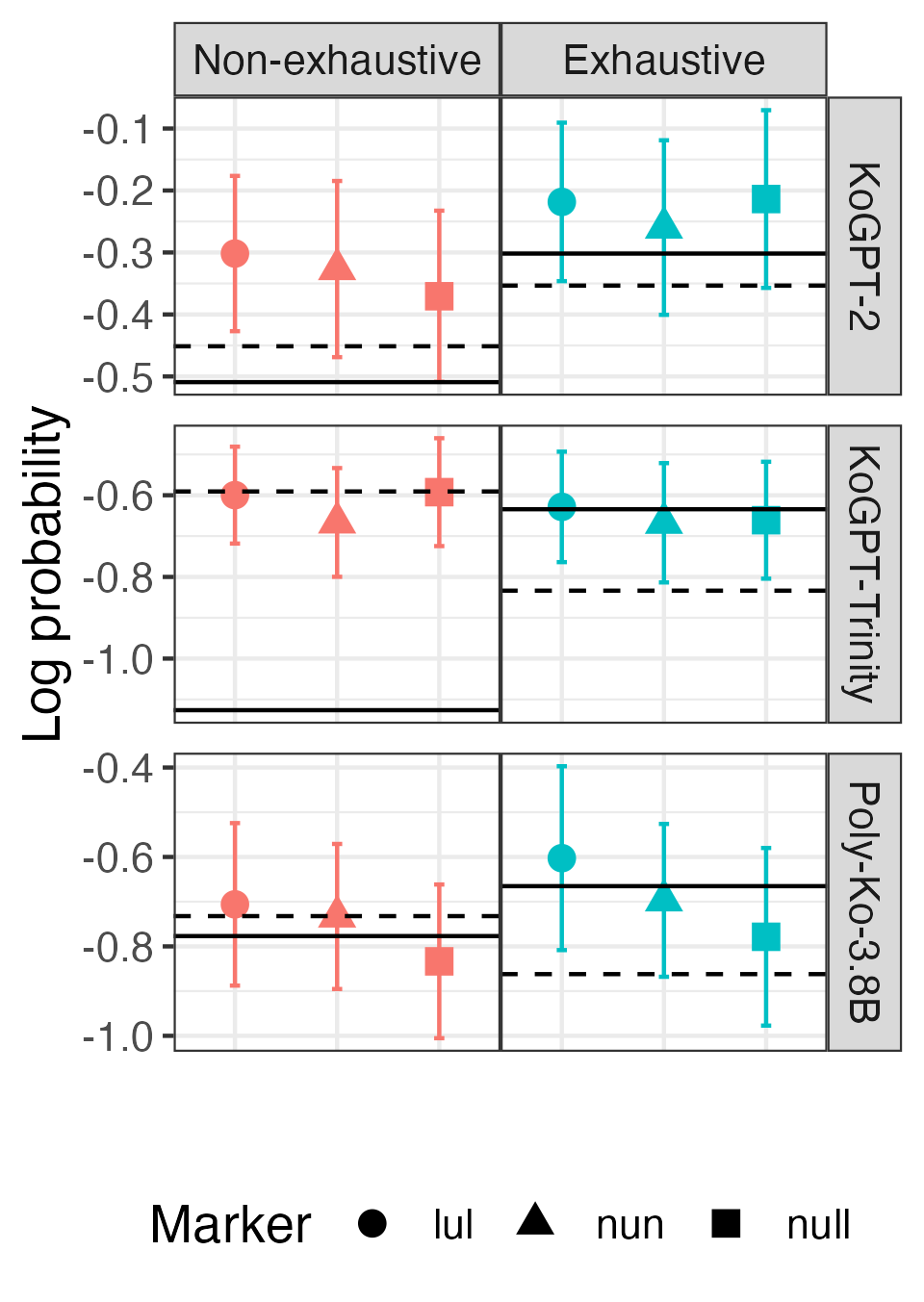}
    \end{subfigure}~ 
    \begin{subfigure}[t]{0.49\textwidth}
        \raggedright        
        \includegraphics[width=0.65\textwidth]{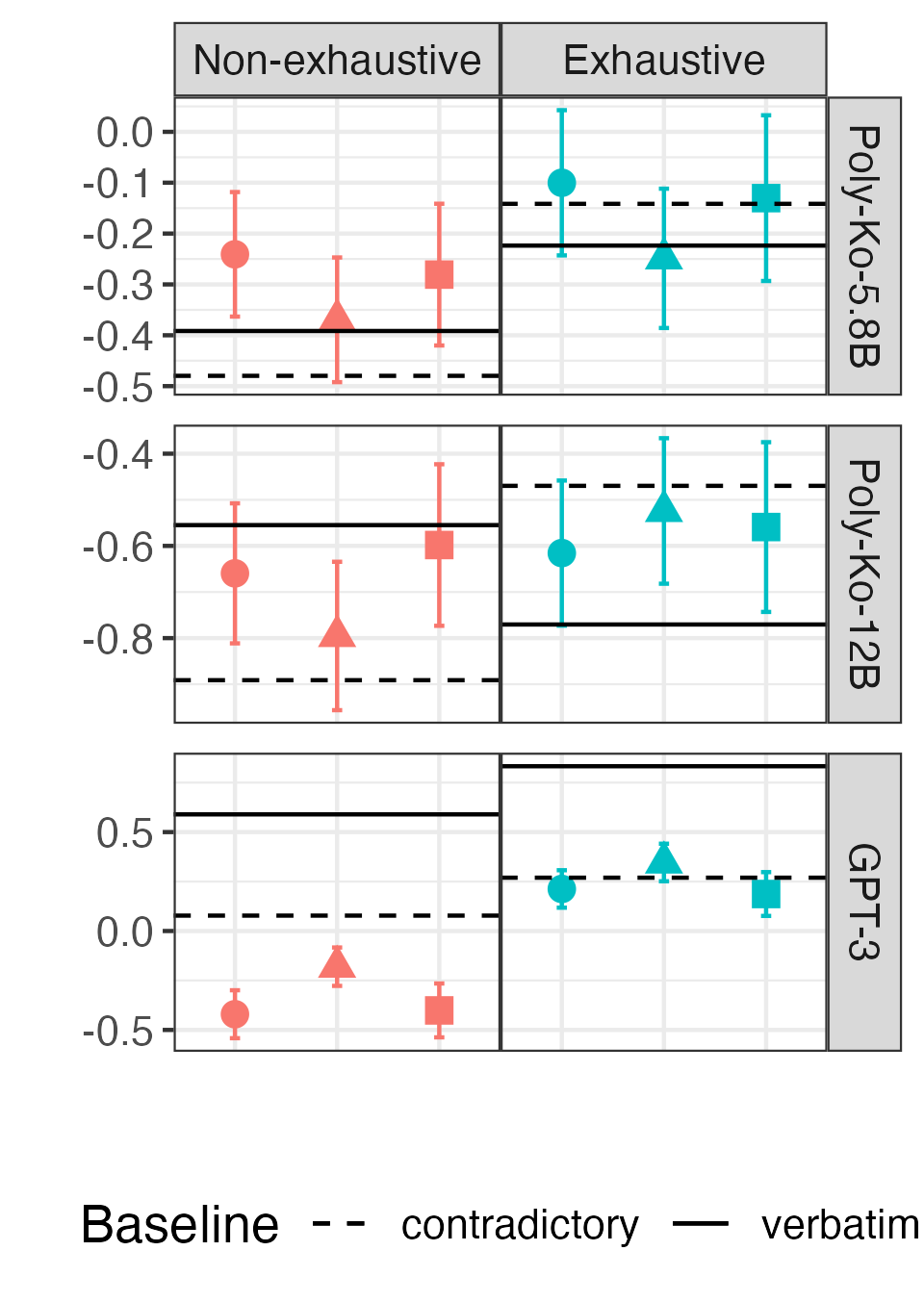}
    \end{subfigure} 
\caption{Mean log probabilities assigned to \textit{lul}-, \textit{nun}-, and null-marked responses when non-exhaustive or exhaustive messages are intended are shown. Error bars indicate 95\% CIs. Dashed horizontal lines indicate the mean log probabilities assigned to contradictory responses, such as ``Received only the medal'' when a non-exhaustive message (both the medal and the trophy) is intended, or ``Received both the trophy and the medal'' when an exhaustive message (only the medal) is intended. Solid horizontal lines indicate the mean log probabilities assigned to verbatim responses, such as ``Received both the trophy and the medal'' when a non-exhaustive message (both the medal and the trophy) is intended, or ``Received only the trophy'' when an exhaustive message (only the medal) is intended.}
\label{fig:production}
\end{figure*}

\subsection{Probability measurements}

\begin{table*}[ht]
    \centering 
    \begin{small}
    \begin{tabular}{l|c|c|c|c|c|c} \hline
         & \multicolumn{6}{c}{$\beta$ of log probability ($p$-value)} \\ \cline{2-7}
         & KoGPT-2 & Ko-Trinity & Poly-3.8B & Poly-5.8B & Poly-12B & GPT-3 \\  \hline
         Intercept & -0.59 (***) & -0.83 (***) & -0.37 (***) & -0.28 (***) & -0.60 (***) & -0.40 (***) \\
         Exhaustivity:Yes & -0.07 (0.14) & 0.05 (0.30) & 0.16 (***) & 0.15 (***) & 0.04 (0.28)& 0.59 (***) \\
         Marker:\textit{lul} &-0.01 (0.88) & 0.13 (*) & 0.07 (0.08) & 0.04 (0.34) & -0.06 (0.09) & -0.02 (0.54) \\ 
         Marker:\textit{nun} &-0.07 (0.11) & 0.10 (0.06) & 0.04 (0.26) & -0.09 (*) & -0.20 (***) & 0.22 (***) \\
         Exhaustivity:Yes$\times$Marker:\textit{lul} & 0.04 (0.55) & 0.05 (0.51) & -0.07 (0.18) & -0.01 (0.87)  & 0.00 (0.92) & 0.04 (0.32) \\          Exhaustivity:Yes$\times$Marker:\textit{nun} & 0.07 (0.31) & -0.02 (0.80) & -0.09 (0.10) & -0.03 (0.62)& \textbf{0.23 (***)} & -0.06 (0.17)  \\  \hline
    \end{tabular}
    \end{small}
    \caption{Log probabilities obtained with the six models are fit with the mixed effects model: \texttt{log probability $\sim$ exhaustivity * marker + (1|item)}. Baseline model is set as Exhaustivity:No, Marker:Null. Positive coefficients indicate that the models assign higher probability to a response in varying exhaustivity in the intended message. ***: $p < 0.001$. *: $p < 0.05$.} 
    \label{tab:production-regression}
\end{table*}

Assessing all models except ChatGPT, we measured log probabilities assigned to the response sentences with different object markings. Log probability of a sentence is obtained in the same manner as in Experiment 1: logits of each token in a sentence are converted into log probabilities, which are then summed and normalized for the number of tokens in a sentence; only with GPT-3, log probabilities are directly accessed.  We report the log probabilities instead of surprisal for this experiment in order to directly reflect the likelihood of generating different object markings.

We created two different types of prompts from one item, one including the intended message statement and one without it---and subtracted the log probabilities obtained from the former from the log probabilities obtained from the latter. This was to show how much the exclusivity status in the intended message is associated with the log probabilities assigned to each response type. In other words, models saw 960 prompts, with 480 containing the intended message statement and 480 without it. We report 480 log probability measurements obtained from each pair from each model.

Results are reported in Figure \ref{fig:production}. Except GPT-3, all models struggled to assign higher probability to verbatim responses than to contradictory responses. Excluding this baseline results, Polyglot-Ko-12B demonstrated some competency in associating the \textit{nun} marker with exhaustivity status: It assigned lower probabilities to \textit{nun}-marked responses when exhaustivity was not intended, and higher probabilities when exhaustivity was intended. 
In the first experiment, we observed that GPT-3 exhibited processing patterns that were partially comparable to humans. However, GPT-3 did not associate the intended exhaustivity provided in the prompt with different marking options, as it did not assign significantly higher probabilities to \textit{lul} or \textit{nun}-marked responses when exhaustivity needed to be delivered.

\subsection{Mixed-effects models}

Probability measurements obtained from the six models in Figure \ref{fig:production} are further analyzed with a set of mixed-effects models. We fitted a mixed-effects model predicting log probability of a response sentence with the intended exhaustivity status of the object and different markings of the object, while controlling random effects of each lexicalization of an item. Results are reported in Table \ref{tab:production-regression}.

Seen with the Polyglot-Ko-3.8B, Polyglot-Ko-5.8B, and GPT-3, larger models are more likely to assign distinct probabilities when intended exhaustivity differs. Although not in the direction that matches patterns in natural language, larger models (Polyglot-Ko-5.8B, Polyglot-Ko-12B, GPT-3) appear at least to gain sensitivity towards paradigmatic selections of marking options, as they assigned significantly different probabilities to \textit{nun}-marked responses. Again, the results with Polyglot-Ko-12B is notable, as it assigns significantly higher probability to \textit{nun}-marked responses when exhaustivity needs to be delivered. 

No models associated intended exhaustivity with the \textit{lul} marker. Since the \textit{lul} marker is canonically a grammatical case marker, this indicates that encoding dual meanings of a marker may be more challenging to LLMs. 
Considering this result with the Experiment 1 (c.f., Figure \ref{fig:processing}), we conclude that significantly different surprisals observed with the \textit{lul} marker in Experiment 1 are unlikely to have come from associating the marker with the exhaustivity implicature. Rather, it is likely that the models' sensitivity towards verb continuations were heightened when more canonical object marker (\textit{lul}) appeared.

\subsection{Forced-choice responses}
We elicited responses from ChatGPT and humans with forced-choice tasks. 
Forced-choice tasks had the \textit{intended message}, the \textit{question} portion (c.f. Table \ref{tab:production_item}), and either a pair of \textit{lul}-marked and null-marked response, or a pair of \textit{nun}-marked and null-marked as `response sets'.
By presenting the options in these pairs, we aimed to ensure that participants and the model did not select the \textit{lul} marker solely based on its grammatical function. If participants or the model associates \textit{lul} and \textit{nun} with exhaustivity implicatures but not with null-markings, the expected choices would be \textit{lul} over null-marked responses and \textit{nun} over null-marked responses.

After providing the response sets, we asked to choose the best message among the two to send as a response to the preceding question.
For ChatGPT, we set the temperature to 0 and provided the system message via OpenAI API: \textit{Please choose the response as you would speak in everyday conversations. Provided options may not express everything that you need to say. Nevertheless, please choose the best option among the two}. 
ChatGPT was presented with one prompt twice, once each with switched order of the response. 
In total, ChatGPT was presented with 192 prompts (48 items with 2 response sets, twice with switched order of response options), all in a zero-shot manner.

\begin{figure}[t]
    \centering
    \includegraphics[width=\columnwidth]{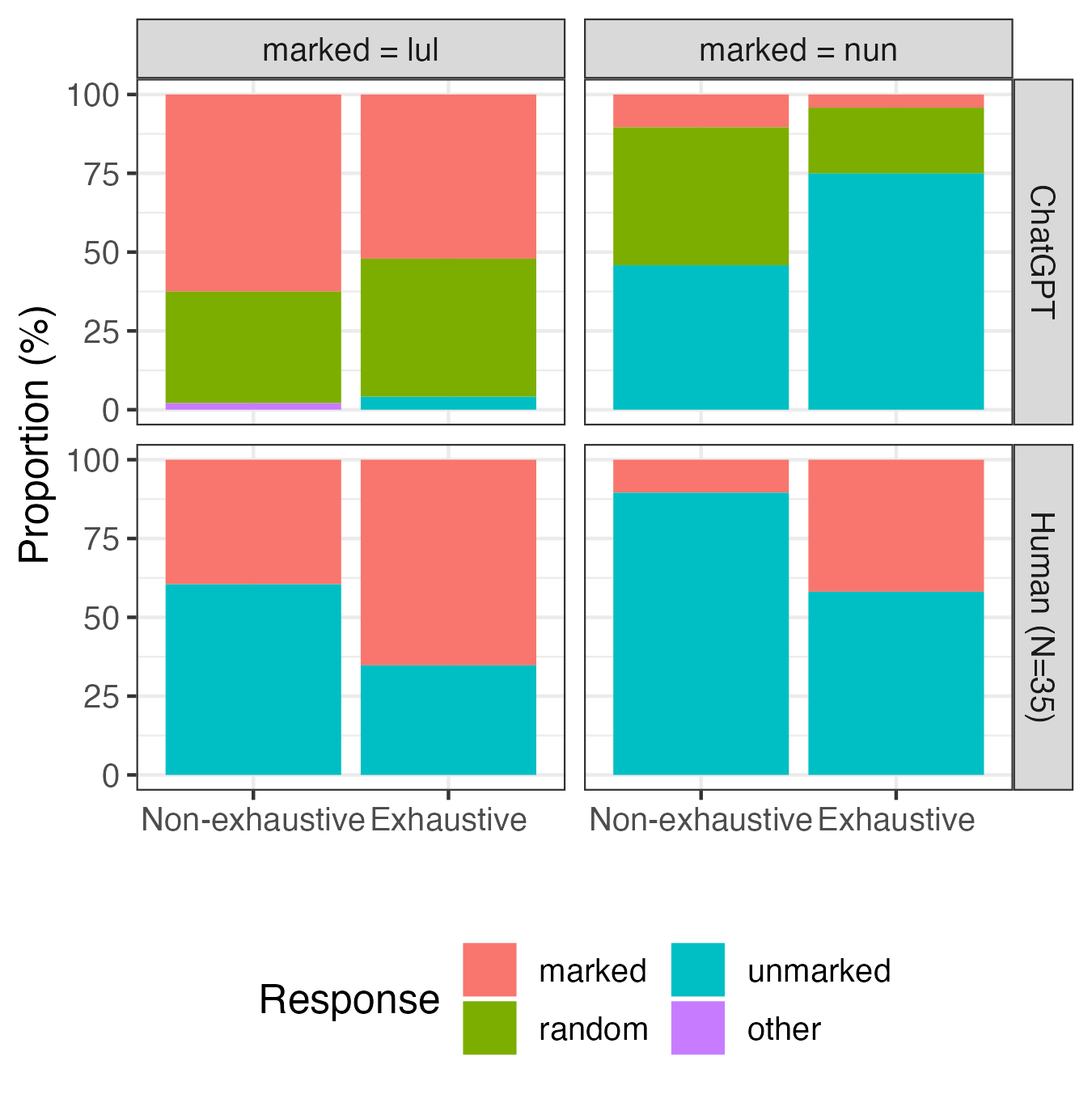}
    \caption{Proportions of responses elicited from ChatGPT and from 35 human participants. The left panels (marked = \textit{lul}) summarize choices when the response sets included \textit{lul}-marked and null-marked objects. Here, the `marked' proportion, colored in red, indicates the proportion of \textit{lul}-marked responses, while the `unmarked' proportion, in blue, indicates the proportion of null-marked responses. On the right panels (marked = \textit{nun}), choices are summarized when response sets included \textit{nun}-marked and null-marked objects. Here, the `marked' proportion, colored in red, indicates the proportion of \textit{nun}-marked responses, while the `unmarked' proportion, in blue, indicates the proportion of null-marked responses.}
    \label{fig:exp2-gpt4human}
\end{figure}

In the human experiment, the order of the marked and unmarked options of responses were randomly switched in every question. Each participant saw 24 critical items, 24 fillers on subject markings, and 4 attention check items. 
Human participants are also given the instruction, after each question, to choose the most proper response even if none of the two can represent everything that the speaker needs to say. 
See the Appendix \ref{sec:appendixC} for the forced-choice tasks written in Korean.
Participants were recruited and compensated via online crowdsourcing platform based in South Korea.\footnote{\url{https://pickply.com/}} After the data elimination process, we report 35 participants' responses.

Results of forced-choice responses are summarized in Figure \ref{fig:exp2-gpt4human}.
To begin, ChatGPT frequduently made `random' choices, meaning that the model didn't select the same option when the order of two responses was flipped. Even when disregarding the random choices, the model exhibited a pattern contrary to that found in human responses. When exhaustivity needed to be expressed, ChatGPT more often selected null-marked responses, whereas humans were more likely to mark the object with either the \textit{lul} or \textit{nun} marker.
Overall, ChatGPT appears incompetent at generating the \textit{lul} or \textit{nun} markers to evoke the implicature in our tasks. 
This result suggests that the decreased ratings observed with repeated continuations in Experiment 1 (c.f., Figure \ref{fig:processing}) are unlikely to stem from the association of the marker with the exhaustivity implicatures.

\section{Discussion}
We observed that some bigger models' results on the \textit{nun} marker were partially coherent with human speakers' patterns.
GPT-3---the largest model among the ones we assessed with log probabilities---showed sensitivity towards the non-cancelable implicature of the \textit{nun} marker (Exp 1), although it did not exhibit sensitivity to the \textit{lul} marker's cancelable exhaustivity implicature (Exp 1, 2). Polyglot-Ko-12B---second to the largest model---exhibited competency in utilizing the \textit{nun} marker to evoke the exhaustivity implicature (Exp 2), but the model didn't show sensitivity towards the implicature's non-cancelability (Exp 1).

The current experiments evaluated LLMs' sensitivity towards discourse pragmatics, not the inferential pragmatics encoded with grammatical structures in a language. Most of the models we tested did not distinguish semantic contradictions from pragmatic infelicity encoded with linguistic structures, nor did they showcase the ability to alternate object markings to generate a particular discourse interpretation. GPT-3, the model trained as an instruction model, and ChatGPT, a model trained to follow conversational instructions, showed patterns closest to how humans associate discourse meanings with the two markers, albeit not identical. Both models demonstrated sensitivity to the discourse interpretation of the discourse marker \textit{nun}, but not to the discourse interpretation of the canonically grammatical case marker \textit{lul}.

In our results, larger-scale models---particularly those fine-tuned using RLHF---produced behavior that was more sensitive to the discourse meanings of morphological markers. This is consistent with past work suggesting that increases in model scale are correlated with improvements in performance \cite{kaplan_scaling_2020}. Of course, many factors were not controlled across the models we tested: the amount of training data, the architecture, whether the model was trained on multiple languages, and more. Future work would benefit from a finer-grained analysis of the corpus data that different models are trained on and examining the impact of the frequency of a given morphological marker with the model's ability to generate behavior sensitive to that marker's implicatures. 

What does this result tell us about the general capabilities of distributional semantics in encoding patterns of natural language? The particular challenge in the current experiments was that distributional semantics needed to encode dual meanings of morphological markers---grammatical object function and exhaustive interpretations established in the context. Distributional semantics, at least as operationalized in LLMs trained without human feedback, do not seem to capture how humans understand the dual functions of the markers. However, providing human feedback and scaling up the embedding space resulted in patterns closer to those in natural language. Thus, encoding dual meanings in multiple domains of language does not appear as an entirely impossible task for distributional semantics to handle.

\section{Conclusion}

The success of Large Language Models in various domains lends support to the hypothesis that much can be learned about the grammatical function and contextual meanings of words from their distributional patterns. In the current work, we examine whether distributional statistics are also sufficient to encode information about the discourse function of words or affixes in addition to their canonical meaning or function. Despite the proven competency of LLMs in grammatical domains, most models tested do not exhibit the human-like ability to use different structures in language to express nuanced meaning in the discourse context. Our study provides baseline assessments of what distributional semantics without any further fine-tuning could achieve.

\section*{Ethics Statement}
All experiments with human participants followed the Institutional Review Board (IRB) guidelines at University of California San Diego. Data and codes are available at \url{https://github.com/hagyeongshin/lm-discourse/}.

\section*{Acknowledgements}
The authors would like to thank Victor Ferreira for advice in human experiments, and the anonymous reviewers for their feedback. Any findings, opinions, and conclusions in this material are those of the authors.

\bibliography{references}

\begin{thebibliography}{29}
\expandafter\ifx\csname natexlab\endcsname\relax\def\natexlab#1{#1}\fi

\bibitem[{Aissen(2003)}]{aissen_differential_2003}
Judith Aissen. 2003.
\newblock Differential {Object} {Marking}: {Iconicity} vs. {Economy}.
\newblock \emph{Natural Language \& Linguistic Theory}, 21:435--483.

\bibitem[{Boleda(2020)}]{boleda_distributional_2020}
Gemma Boleda. 2020.
\newblock \href {https://doi.org/10.1146/annurev-linguistics-011619-030303} {Distributional {Semantics} and {Linguistic} {Theory}}.
\newblock \emph{Annual Review of Linguistics}, 6(1):213--234.

\bibitem[{Bossong(1991)}]{wanner_differential_1991}
Georg Bossong. 1991.
\newblock \href {https://doi.org/10.1075/cilt.69.14bos} {Differential {Object} {Marking} in {Romance} and {Beyond}}.
\newblock In Dieter Wanner and Douglas~A. Kibbee, editors, \emph{New analyses in {Romance} linguistics}, pages 143--170. John Benjamins Publishing Company, Amsterdam.

\bibitem[{Büring(2003)}]{buring_d-trees_2003}
Daniel Büring. 2003.
\newblock On {D}-{Trees}, {Beans}, {And} {B}-{Accents}.
\newblock \emph{Linguistics and Philosophy}, 26:511--545.

\bibitem[{Choi(1996)}]{choi_optimizing_1996}
Hye-Won Choi. 1996.
\newblock \emph{Optimizing structure in context: {Scrambling} and information structure}.
\newblock Ph.D. thesis, Stanford University.

\bibitem[{Choi-Jonin(2008)}]{choi-jonin_particles_2008}
Injoo Choi-Jonin. 2008.
\newblock Particles and {Propositions} in {Korean}.
\newblock In \emph{Adpositions: {Pragmatics}, {Semantic} and {Syntactic} {Perspectives}}, number~74 in Typological {Studies} in {Language}, pages 133--170. John Benjamins Publishing Company, Amsterdam/Philadelphia.

\bibitem[{Futrell et~al.(2019)Futrell, Wilcox, Morita, Qian, Ballesteros, and Levy}]{futrell_neural_2019}
Richard Futrell, Ethan Wilcox, Takashi Morita, Peng Qian, Miguel Ballesteros, and Roger Levy. 2019.
\newblock Neural language models as psycholinguistic subjects: {Representations} of syntactic state.
\newblock In \emph{Proceedings of the 2019 {Conference} of the {North} {American} {Chapter} of the {Association} for {Computational} {Linguistics}: {Human} {Language} {Technologies}}, pages 32--42, Minneapolis, Minnesota. Association for Computational Linguistics.

\bibitem[{Grice(1989)}]{grice_studies_1989}
H.~Paul Grice. 1989.
\newblock \emph{Studies in the {Way} of {Words}}.
\newblock Harvard University Press, Cambridge, MA.

\bibitem[{Harris(1954)}]{harris_distributional_1954}
Zellig~S. Harris. 1954.
\newblock \href {https://doi.org/10.1080/00437956.1954.11659520} {Distributional {Structure}}.
\newblock \emph{\textit{WORD}}, 10(2-3):146--162.

\bibitem[{Horn(1981)}]{horn_exhaustiveness_1981}
Laurence~R Horn. 1981.
\newblock Exhaustiveness and the {Semantics} of {Clefts}.
\newblock In \emph{Proceedings of the {Eleventh} {Annual} {Meeting} of the {North} {Eastern} {Linguistic} {Society}}, volume~11, pages 125--142.

\bibitem[{Hu et~al.(2023)Hu, Floyd, Jouravlev, Fedorenko, and Gibson}]{hu_fine-grained_2023}
Jennifer Hu, Sammy Floyd, Olessia Jouravlev, Evelina Fedorenko, and Edward Gibson. 2023.
\newblock A fine-grained comparison of pragmatic language understanding in humans and language models.
\newblock In \emph{Proceedings of the 61st {Annual} {Meeting} of the {Association} for {Computational} {Linguistics} ({Volume} 1: {Long} {Papers})}, pages 4194--4213, Toronto, Canada. Association for Computational Linguistics.

\bibitem[{Hu and Levy(2023)}]{hu_prompt-based_2023}
Jennifer Hu and Roger Levy. 2023.
\newblock Prompt-based methods may underestimate large language models' linguistic generalizations.
\newblock In \emph{Proceedings of the 2023 {Conference} on {Empirical} {Methods} in {Natural} {Language} {Processing}}, pages 5040--5060. Association for Computational Linguistics.

\bibitem[{Jeretič et~al.(2020)Jeretič, Warstadt, Bhooshan, and Williams}]{jeretic_are_2020}
Paloma Jeretič, Alex Warstadt, Suvrat Bhooshan, and Adina Williams. 2020.
\newblock Are {Natural} {Language} {Inference} {Models} {IMPPRESsive}? {Learning} {IMPlicature} and {PRESupposition}.
\newblock In \emph{Proceedings of the 58th {Annual} {Meeting} of the {Association} for {Computational} {Linguistics}}, pages 8690--8705, Online. Association for Computational Linguistics.

\bibitem[{Kaplan et~al.(2020)Kaplan, McCandlish, Henighan, Brown, Chess, Child, Gray, Radford, Wu, and Amodei}]{kaplan_scaling_2020}
Jared Kaplan, Sam McCandlish, Tom Henighan, Tom~B. Brown, Benjamin Chess, Rewon Child, Scott Gray, Alec Radford, Jeffrey Wu, and Dario Amodei. 2020.
\newblock \href {http://arxiv.org/abs/2001.08361} {Scaling {Laws} for {Neural} {Language} {Models}}.
\newblock ArXiv:2001.08361 [cs, stat].

\bibitem[{Kim(2018)}]{kim_deriving_2018}
Jieun Kim. 2018.
\newblock \href {https://doi.org/10.1007/s10988-017-9227-6} {Deriving the contrastiveness of contrastive -nun in {Korean}}.
\newblock \emph{Linguistics and Philosophy}, 41(4):457--482.

\bibitem[{Kwon and Zribi-Hertz(2008)}]{kwon_differential_2008}
Song-Nim Kwon and Anne Zribi-Hertz. 2008.
\newblock Differential {Function} {Marking}, {Case}, and {Information} {Structure}: {Evidence} from {Korean}.
\newblock \emph{Language}, 84(2):258--299.

\bibitem[{Lambrecht(1994)}]{lambrecht_information_1994}
Knud Lambrecht. 1994.
\newblock \emph{Information {Structure} and {Sentence} {Form}: {Topic}, focus, and the mental representations of discourse referents}.
\newblock Cambridge University Press.

\bibitem[{Lee(2003)}]{lee_contrastive_2003}
Chungmin Lee. 2003.
\newblock Contrastive topic and/or contrastive focus.
\newblock In \emph{Japanese/{Korean} {Linguistics}}, volume~12, pages 352--364.

\bibitem[{Lee(2017)}]{lee_contrastiveness_2017}
Chungmin Lee. 2017.
\newblock \href {https://doi.org/10.1007/978-3-319-10106-4} {\emph{Contrastiveness in {Information} {Structure}, {Alternatives} and {Scalar} {Implicatures}}}.
\newblock 91. Springer International Publishing, Cham, Switzerland.

\bibitem[{Lee(2006)}]{lee_iconicity_2006}
Hanjung Lee. 2006.
\newblock Iconicity and {Variation} in the {Choice} of {Object} {Forms} in {Korean}.
\newblock \emph{Language Research}, 42(2):323--355.

\bibitem[{Lee(2007)}]{lee_case_2007}
Hanjung Lee. 2007.
\newblock \href {https://doi.org/10.1016/j.pragma.2007.04.012} {Case ellipsis at the grammar/pragmatics interface: {A} formal analysis from a typological perspective}.
\newblock \emph{Journal of Pragmatics}, 39(9):1465--1481.

\bibitem[{Lenci(2018)}]{lenci_distributional_2018}
Alessandro Lenci. 2018.
\newblock \href {https://doi.org/10.1146/annurev-linguistics-030514-125254} {Distributional {Models} of {Word} {Meaning}}.
\newblock \emph{Annual Review of Linguistics}, 4(1):151--171.

\bibitem[{Marelli and Baroni(2015)}]{marelli_affixation_2015}
Marco Marelli and Marco Baroni. 2015.
\newblock \href {https://doi.org/10.1037/a0039267} {Affixation in semantic space: {Modeling} morpheme meanings with compositional distributional semantics.}
\newblock \emph{Psychological Review}, 122(3):485--515.

\bibitem[{Rooth(1992)}]{rooth_theory_1992}
Mats Rooth. 1992.
\newblock \href {https://doi.org/10.1007/BF02342617} {A theory of focus interpretation}.
\newblock \emph{Natural Language Semantics}, 1(1):75--116.

\bibitem[{Tenney et~al.(2019{\natexlab{a}})Tenney, Das, and Pavlick}]{tenney_bert_2019}
Ian Tenney, Dipanjan Das, and Ellie Pavlick. 2019{\natexlab{a}}.
\newblock {BERT} {Rediscovers} the {Classical} {NLP} {Pipeline}.
\newblock In \emph{Proceedings of the 57th {Annual} {Meeting} of the {Association} for {Computational} {Linguistics}}, pages 4593--4601. Association for Computational Linguistics.

\bibitem[{Tenney et~al.(2019{\natexlab{b}})Tenney, Xia, Chen, Wang, Poliak, McCoy, Kim, Van~Durme, Bowman, Das, and Pavlick}]{tenney_what_2019}
Ian Tenney, Patrick Xia, Berlin Chen, Alex Wang, Adam Poliak, R.~Thomas McCoy, Najoung Kim, Benjamin Van~Durme, Samuel~R. Bowman, Dipanjan Das, and Ellie Pavlick. 2019{\natexlab{b}}.
\newblock What do you learn from context? {Probing} for sentence structure in contextualized word representations.

\bibitem[{Trott and Bergen(2021)}]{trott_raw-c_2021}
Sean Trott and Benjamin Bergen. 2021.
\newblock {RAW}-{C}: {Relatedness} of {Ambiguous} {Words}--in {Context} ({A} {New} {Lexical} {Resource} for {English}).
\newblock In \emph{Proceedings of the 59th {Annual} {Meeting} of the {Association} for {Computational} {Linguistics} and the 11th {International} {Joint} {Conference} on {Natural} {Language} {Processing} ({Volume} 1: {Long} {Papers})}, pages 7077--7087, Online. Association for Computational Linguistics.

\bibitem[{Trott et~al.(2023)Trott, Jones, Chang, Michaelov, and Bergen}]{trott_large_2023}
Sean Trott, Cameron Jones, Tyler Chang, James Michaelov, and Benjamin Bergen. 2023.
\newblock \href {https://doi.org/10.1111/cogs.13309} {Do {Large} {Language} {Models} {Know} {What} {Humans} {Know}?}
\newblock \emph{Cognitive Science}, 47(7):e13309.

\bibitem[{Van~Rooij and Schulz(2004)}]{van_rooij_exhaustive_2004}
Robert Van~Rooij and Katrin Schulz. 2004.
\newblock \href {https://doi.org/10.1007/s10849-004-2118-6} {Exhaustive {Interpretation} of {Complex} {Sentences}}.
\newblock \emph{Journal of Logic, Language and Information}, 13(4):491--519.

\end{thebibliography}
\bibliographystyle{acl_natbib}

\appendix

\section{Items used in Experiment 1}\label{sec:appendixA}

An example of items provided to LLMs excluding ChatGPT. 
\vspace{-1em}

\noindent \begin{mdframed}

소희는 미나와 유나가 메달과 트로피를 받을 수 있었다는 것을 알고 있습니다.
\vspace{.5em}

\noindent 소희: 유나는 뭘 받았어? \\
\noindent 미나: \{메달을 받았어./메달은 받았어.\} \{트로피도 받았어./트로피는 못 받았어./트로피만 받았어.\}
\end{mdframed}

\vspace{2em}

\noindent An example of items provided to ChatGPT, with the system message set as ``반드시 1과 7사이의 숫자 중 하나로만 답해 주세요.''
\vspace{-1em}

\noindent
\begin{mdframed}

소희는 미나와 유나가 메달과 트로피를 받을 수 있었다는 것을 알고 있습니다.
\vspace{.5em}

\noindent 소희: 유나는 뭘 받았어?\\
\noindent 미나: \{메달을 받았어./메달은 받았어.\} \{트로피도 받았어./트로피는 못 받았어./트로피만 받았어.\}
\vspace{.5em}

\noindent 위의 대화에서 미나가 첫 번째 답장에 이어 두 번째 답장을 말한 것이 얼마나 적절한지 1과 7 사이의 숫자로 답해 주세요. 1은 `전혀 적절하지 않다'는 것을 뜻하고 7은 `매우 적절하다'는 것을 뜻합니다.
\end{mdframed}

\vspace{1em}

\noindent An example of items provided to human participants.
\vspace{-1em}

\noindent 
\begin{mdframed}

소희, 미나, 유나가 함께 마라톤 경주에 참여하기로 했습니다. 마라톤 경주에 참여한 사람들은 메달과 트로피를 받을 수 있었습니다. 미나와 유나는 약속한 대로 함께 경주에 참여했습니다. 소희는 약속을 지키지 않았기 때문에 경주에서 누가 무엇을 받았는지 모르고 있습니다. 이후 소희와 미나는 다음 페이지에 나와 있는 문자메시지를 주고 받았습니다.

\vspace{.5em}

\hrule

\vspace{.5em}

\noindent
아래에 소희와 미나가 마라톤 경주에 대해 이야기한 문자 메시지가 주어져 있습니다.
\vspace{.5em}

\noindent 소희: 유나는 뭘 받았어?\\
\noindent 미나: \{메달을 받았어./메달은 받았어.\}
\{트로피도 받았어./트로피는 못 받았어.\}

\vspace{.5em}

\noindent 위의 대화에서 미나가 첫 번째 답장에 이어 두 번째 답장을 말한 것이 얼마나 적절하다고 생각하십니까?
\vspace{.5em}

\noindent 전혀 적절하지 않다 \hfill 매우 적절하다

\hspace{.3em} 1 \hspace{1.5em} 2 \hspace{1.5em} 3 \hspace{1.5em} 4 \hspace{1.5em} 5 \hspace{1.5em} 6 \hspace{1.5em} 7

\end{mdframed}

\begin{figure}[h!]
     \centering
     \begin{subfigure}[t]{0.45\columnwidth}
         \centering
         \includegraphics[width=\columnwidth]{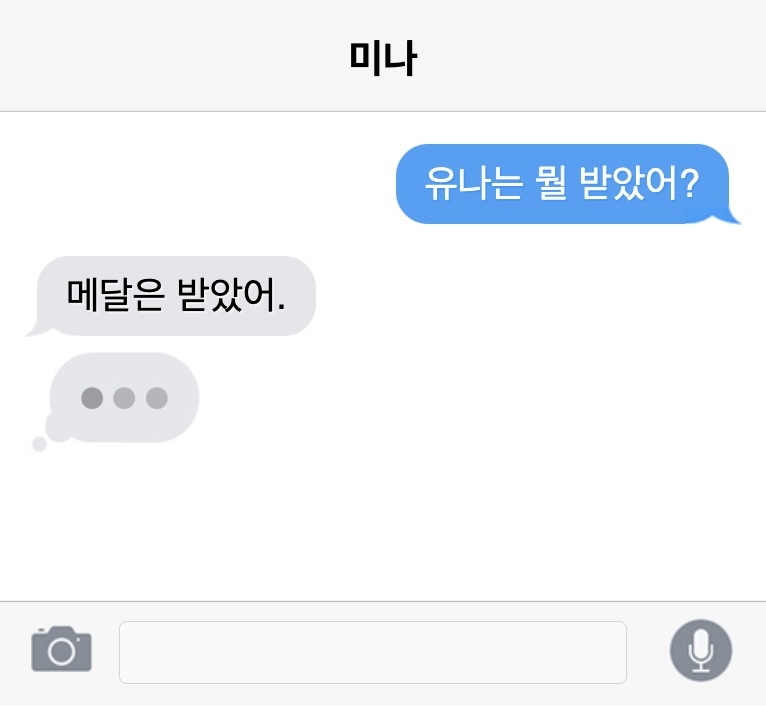}
         \label{fig:kor-item-p1}
     \end{subfigure} $\rightarrow$
     \begin{subfigure}[t]{0.45\columnwidth}
         \centering
         \includegraphics[width=\columnwidth]{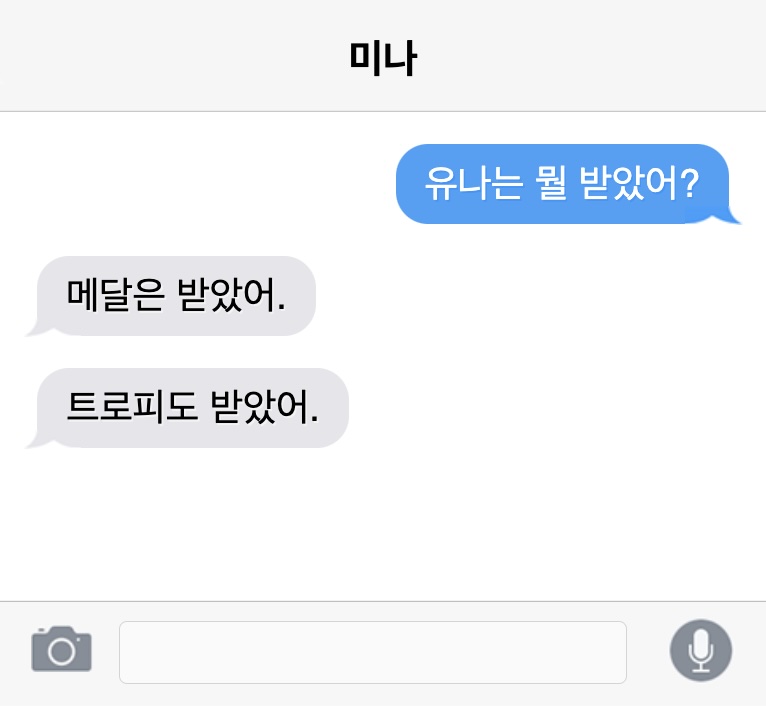}
         \label{fig:kor-item-p2}
     \end{subfigure}
        \caption{The question and answer portions in the human experiment items were presented in an interface resembling that of mobile text messages. Each message appeared in a 3-second interval within a short video clip.}
        \label{fig:three graphs}
\end{figure}

\section{Raw ratings from Experiment 1}\label{sec:appendixB}
\begin{figure}[h!]
    \centering
    \includegraphics[width=0.9\columnwidth]{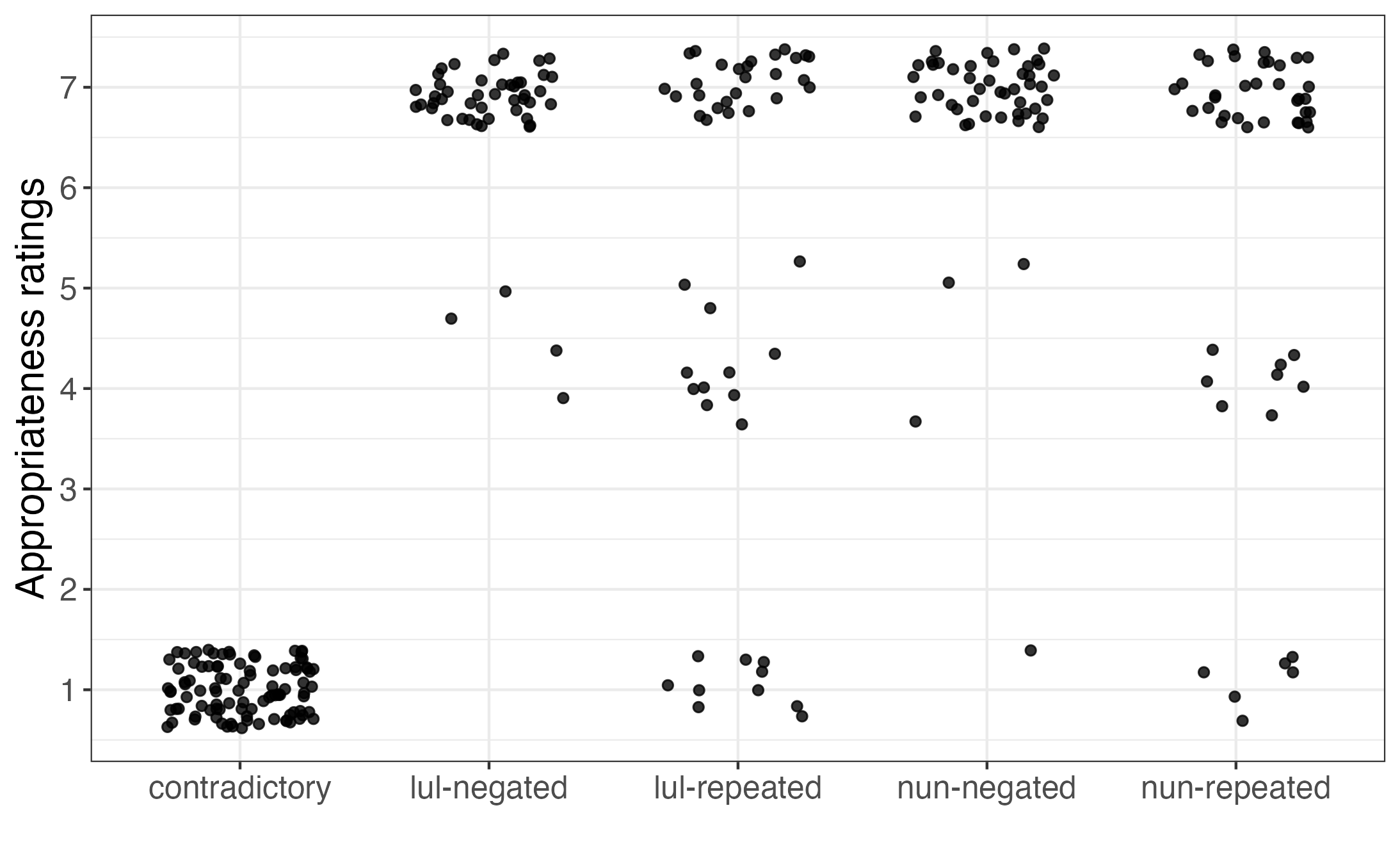}
    \caption{Raw ratings obtained from ChatGPT in Experiment 1 (1 = not approriate at all, 7 = highly appropriate).}
    \label{fig:enter-label}
\end{figure}

\begin{figure}[h!]
    \centering
    \includegraphics[width=0.9\columnwidth]{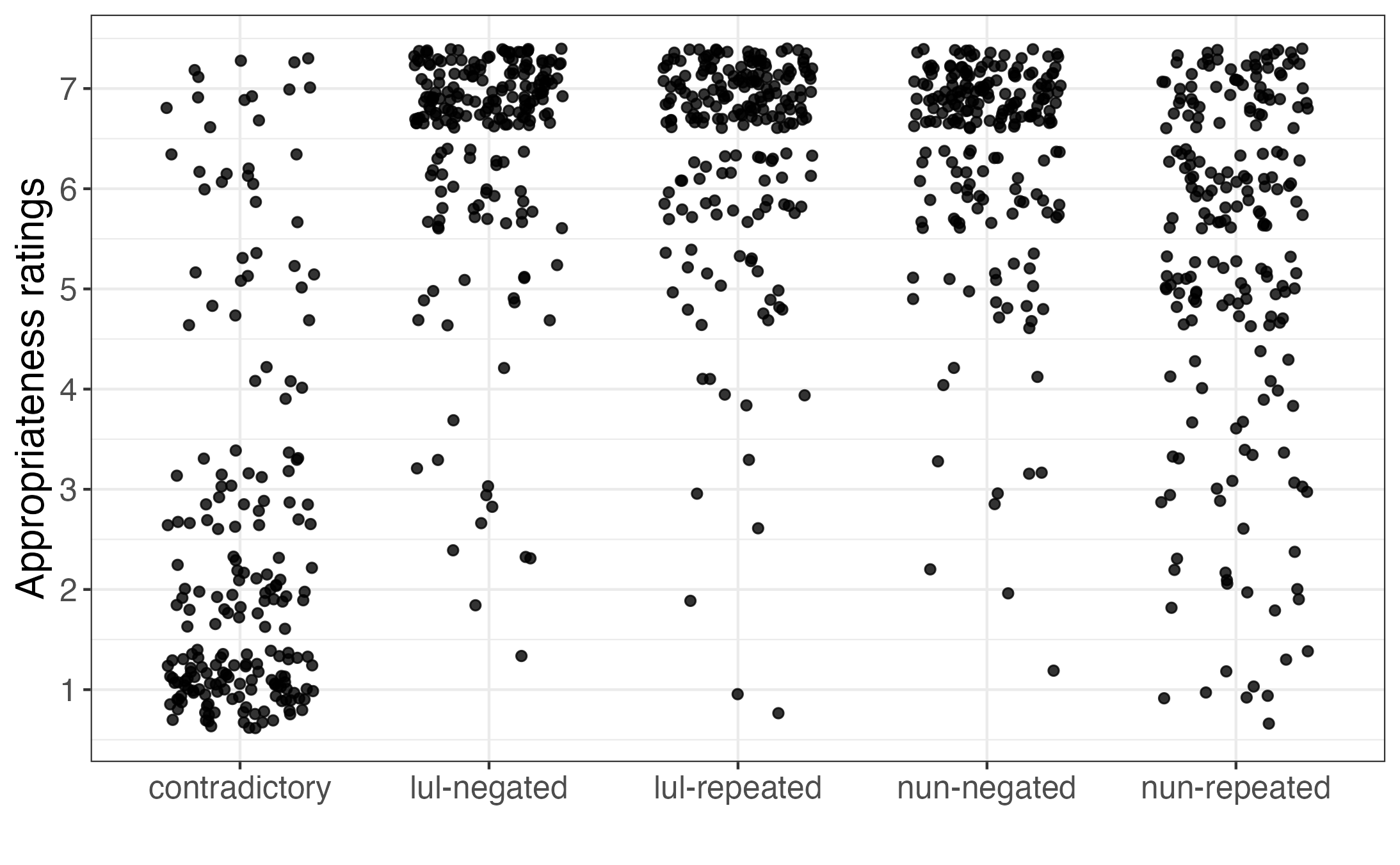}
    \caption{Raw ratings obtained from 34 human participants in Experiment 1 (1 = not approriate at all, 7 = highly appropriate).}
    \label{fig:enter-label}
\end{figure}

\section{Items used in Experiment 2}\label{sec:appendixC}

An example of items provided to LLMs, excluding ChatGPT.

\vspace{-1em}

\noindent \begin{mdframed}

\{미나는 유나가 메달만 땄다고 답하려고 합니다./미나는 유나가 메달과 트로피를 둘 다 땄다고 답하려고 합니다.\} 

\vspace{0.5em}

\noindent 소희: 유나는 뭘 받았어? \\
\noindent 미나: \{메달을 받았어./메달 받았어./메달은 받았어./메달만 받았어./메달이랑 트로피를 둘 다 받았어.\}
\end{mdframed}

\vspace{1em}

\noindent An example of items provided to ChatGPT, with the system message set as ``일상적인 대화 상황을 생각하고 답장을 골라 주세요. 주어지는 답장들은 말해야 하는 모든 것을 나타내지 않을 수 있습니다. 이와 상관 없이 두 개의 답장 중에 더 적절하다고 생각하는 답장을 골라 주세요.''

\vspace{-1em}

\noindent \begin{mdframed}

\{미나는 유나가 메달만 땄다고 답하려고 합니다./미나는 유나가 메달과 트로피를 둘 다 땄다고 답하려고 합니다.\} 

\vspace{0.5em}

\noindent 소희: 유나는 뭘 받았어? \\
\noindent 미나: \{메달을 받았어. 메달 받았어./메달은 받았어. 메달 받았어.\}

\end{mdframed}

\vspace{1em}

\noindent An example of items provided to human participants.

\vspace{-1em}
\noindent \begin{mdframed}
소희, 미나, 유나가 함께 마라톤 경주에 참여하기로 했습니다. 마라톤 경주에는 메달과 트로피이 걸려 있었습니다. 미나와 유나는 약속한 대로 함께 경주에 참여했습니다. 미나는 유나가 메달을 땄고 트로피를 따지 못했다는 것을 알고 있습니다. 소희는 약속을 지키지 않았기 때문에 유나가 무엇을 땄는지 모르고 있습니다. 이후 소희는 미나에게 아래에 나와 있는 문자메시지를 보내왔습니다.	
\vspace{0.5em}

\noindent 소희: 유나는 뭘 받았어?

\vspace{0.5em}

\noindent 미나는 유나가 메달만 땄다고 답하려고 합니다. 아래에 주어진 문장들 중 미나가 답장으로 보내기에 가장 적절한 것은 무엇입니까? 아래에 주어진 문장들은 미나가 말해야 하는 모든 것을 나타내지 않을 수 있습니다. 이와 상관 없이 미나가 보내기에 최선이라고 생각하는 답장을 선택해 주세요. 

\vspace{0.5em}

\noindent 선택지1: $\bigcirc$ 메달을 받았어. $\bigcirc$  메달 받았어.\\
선택지2: $\bigcirc$ 메달은 받았어. $\bigcirc$ 메달 받았어.
\end{mdframed}

\begin{figure}[h]
    \centering
\includegraphics[width=0.9\columnwidth]{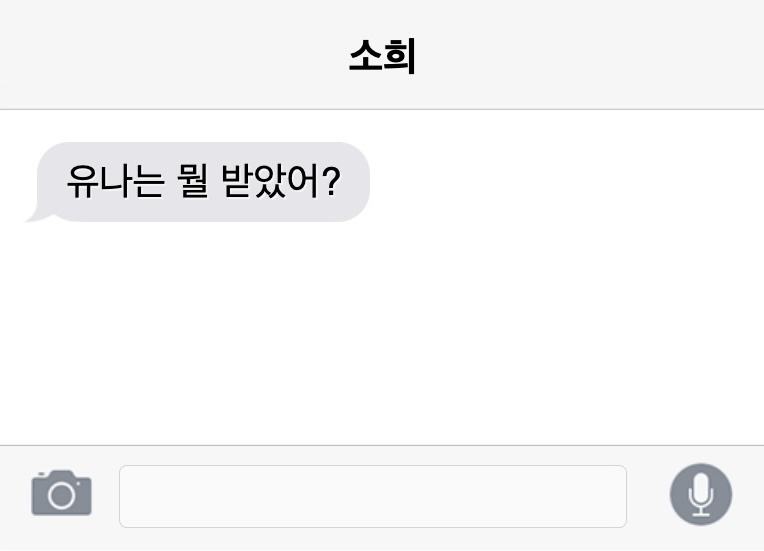}
\includegraphics[width=0.45\columnwidth]{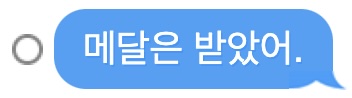}
\includegraphics[width=0.45\columnwidth]{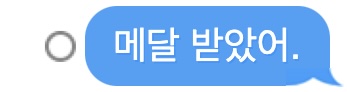}
    \caption{The question and response options in the human experiment items were presented in an interface resembling that of mobile text messages.}
    \label{fig:enter-label}
\end{figure}

\end{document}